\documentclass[sigconf]{acmart}

\usepackage{amsmath,amsfonts}
\usepackage{subcaption}
\usepackage{graphicx}
\usepackage{multirow}
\usepackage{enumitem}

\AtBeginDocument{%
  \providecommand\BibTeX{{%
    \normalfont B\kern-0.5em{\scshape i\kern-0.25em b}\kern-0.8em\TeX}}}

\setcopyright{acmlicensed}
\copyrightyear{2025}
\acmYear{2025}
\acmDOI{XXXXXXX.XXXXXXX}

\acmConference[CIKM'25]{conference details}{\textcolor{red}{Under Review} 
  }
  {
  }
\acmISBN{XXXXX}

\begin{document}

\title{Challenging Gradient Boosted Decision Trees with Tabular Transformers for Fraud Detection at Booking.com}

\author{Sergei Krutikov}
\authornote{Equal contribution}
\authornote{The author no longer works for Booking.com, but this work was conducted during his full-time employment at Booking.com}
\email{drauggil@gmail.com}
\affiliation{%
  \institution{Booking.com}
  \city{Amsterdam}
  \country{the Netherlands}
}

\author{Bulat Khaertdinov}
\authornotemark[1] 
\authornote{This work was conducted during an internship at Booking.com} 
\email{b.khaertdinov@maastrichtuniversity.nl}
\orcid{}
\affiliation{%
  \institution{Maastricht University}
  \city{Maastricht}
  \country{the Netherlands}
}

\author{Rodion Kiriukhin}
\email{rodion.kiriukhin@booking.com}
\affiliation{%
  \institution{Booking.com}
  \city{Amsterdam}
  \country{the Netherlands}
}

\author{Shubham Agrawal}
\email{shubham.agrawal@booking.com}
\affiliation{%
  \institution{Booking.com}
  \city{Amsterdam}
  \country{the Netherlands}
}

\author{Mozhdeh Ariannezhad}
\email{mozhdeh.ariannezhad@booking.com}
\affiliation{%
  \institution{Booking.com}
  \city{Amsterdam}
  \country{the Netherlands}
}

\author{Kees Jan De Vries}
\email{kees.devries@booking.com}
\affiliation{%
  \institution{Booking.com}
  \city{Amsterdam}
  \country{the Netherlands}
}

\renewcommand{\shortauthors}{Krutikov and Khaertdinov, et al.}

\begin{abstract}
Transformer-based neural networks, empowered by Self-Supervised Learning (SSL), have demonstrated unprecedented performance across various domains. However, related literature suggests that tabular Transformers may struggle to outperform classical Machine Learning algorithms, such as Gradient Boosted Decision Trees (GBDT). In this paper, we aim to challenge GBDTs with tabular Transformers on a typical task faced in e-commerce, namely fraud detection. Our study is additionally motivated by the problem of selection bias, often occurring in real-life fraud detection systems. It is caused by the production system affecting which subset of traffic becomes labeled. This issue is typically addressed by sampling randomly a small part of the whole production data, referred to as a Control Group. This subset follows a target distribution of production data and therefore is usually preferred for training classification models with standard ML algorithms. Our methodology leverages the capabilities of Transformers to learn transferable representations using all available data by means of SSL, giving it an advantage over classical methods. Furthermore, we conduct large-scale experiments, pre-training tabular Transformers on vast amounts of data instances and fine-tuning them on smaller target datasets. The proposed approach outperforms heavily tuned GBDTs by a considerable margin of the Average Precision (AP) score in offline evaluations. Finally, we report the results of an online A/B experiment. Experimental results confirm the superiority of tabular Transformers compared to GBDTs in production, demonstrated by a statistically significant improvement in our business metric.
 
\end{abstract}






\maketitle

\section{Introduction}
\label{sec:intro}
Deep Learning has achieved tremendous success in various applications that deal with unstructured data, such as Computer Vision, Speech Recognition, Natural Language and Signal Processing. Moreover, in the past decade, Deep Learning methods employed in these fields have developed from feed-forward neural networks with thousands of parameters to large and sophisticated architectures with billions of parameters based on such techniques as attention-based Transformers~\cite{vaswani2017attention}. These state-of-the-art models can learn complex patterns in data, effectively capturing properties of and dependencies between input features, and encode it into meaningful representations even without using labeled data via Self-Supervised Learning (SSL)~\cite{balestriero2023cookbook}. 


A number of recent studies that have adopted these advancements to structured tabular data include both Transformers~\cite{huang2020tabtransformer, gorishniy2021revisiting} and SSL~\cite{zhang2020vime, bahri2021scarf, rubachev2022revisiting}. However, they show that these methods do not always outperform powerful classical Machine Learning algorithms, such as Gradient Boosted Decision Trees (GBDTs) \cite{shwartz2022tabular}. Some works on tabular representation learning argue that this could be due to certain properties of tabular data. Specifically, it is an unordered set of heterogeneous features represented in different formats (categorical, numerical, ordinal) that do not contain spatial or sequential patterns~\cite{somepalli2021saint}. In contrast, many Deep Learning methods are particularly tailored to extract features from spatial or sequential structures and learning dependencies between homogeneous inputs, e.g. pixel values in images or (sub-)words in a sentence. 

Nonetheless, related literature suggests that GBDTs lack the ability to transfer knowledge and extract information from unlabeled data \cite{levin2023transfer}. The former enables the knowledge sharing between domains, tasks, and datasets, and is typically based on learning transferable representations of the input data. The latter exposes general patterns in unlabeled data and, in this way, improves downstream task performance. This is also known as semi-supervised learning and can be implemented through the "pre-train and fine-tune" paradigm that has become very popular due to SSL. Both these capabilities are especially useful for low-resource settings when limited amounts of labeled data are available or when data annotation is costly. Such settings are typical for business problems that contain humans in the loop. 

In this paper, we adopt recent advances in tabular Deep Learning to such a problem, namely fraud detection, at scale. Detecting fraud is one of the fundamental problems that has to be addressed by any online platform, including Booking.com. Fraud detection can be framed as a binary classification problem: a discriminative model predicts whether a certain user (e.g. guest or partner) action on a platform is fraudulent using a set of heterogeneous features describing that event. However, various challenges make it a much more difficult task in a real-world industrial environment. One of them is a severe data imbalance as fraudulent actions happen much rarer than genuine ones. Another difficulty is the struggle to obtain a labeled dataset that is not subject to \textit{selection bias}. 
The source of selection bias in fraud detection systems varies based on the characteristics of a certain domain. Some domains require manual investigation by fraud analysts, who do not have enough capacity to annotate all incoming traffic and are typically focused on high-risk cases, producing labels only for the selected subsets. In other domains, a fraud detection system introduces the bias itself as ground truth labels for blocked user actions cannot be restored reliably. In that case, the set of remaining instances cannot be effectively used for model training as its distribution is biased w.r.t.~the true population~\cite{tax2021bookingfraud}.

\begin{figure}[!t]
    \includegraphics[width=0.9\linewidth]{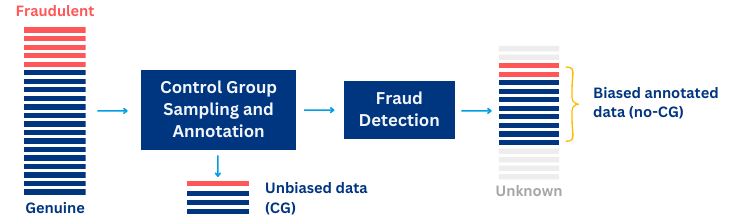}
    \centering
    \vspace{-4pt}
    \caption{Visualization of the instance flow with Control Group sampling strategy.}
    \label{fig:CG}
\end{figure}

These challenges can be mitigated using a \textbf{Control Group (CG)} strategy shown in Figure \ref{fig:CG}, where a small, randomly sampled fraction of all instances from the whole traffic is annotated. Such a strategy helps to obtain a labeled sample of the population without selection bias, but it incurs a significant extra cost associated with annotating such a dataset. What is more, fraud detection models are often trained only on this small part of all data, mostly ignoring vast amounts of other data points. In this paper, we aim to address these challenges by exploring various pre-training strategies for tabular Transformers. The main contributions of our work are listed below:

\vspace{-2pt}
\begin{itemize}[leftmargin=*]
     \item We adapt tabular Transformers, namely an FT-Transformer architecture~\cite{gorishniy2021revisiting}, to the problem of fraud detection.
     \item We propose leveraging data outside of the Control Group for pre-training an FT-Transformer backbone.
    More precisely, we utilize large biased datasets, consisting of tens of millions of instances, for pre-training whereas smaller CG-only data is used for model fine-tuning. 
    \item We investigate how pre-training affects the performance of FT-Transformer and its GBDT competitor in extensive offline experiments.
    We demonstrate that SSL pre-training consistently outperforms training a supervised model from scratch. 
    \item We report the results of an online A/B experiment comparing the performance of FT-Transformer and its GBDT competitor in production. We observe statistically significant improvements in terms of our key business metric.
\end{itemize}

\vspace{-14pt}
\section{Related Work}

\begin{figure*}[!t]
    \includegraphics[width=0.6\linewidth]{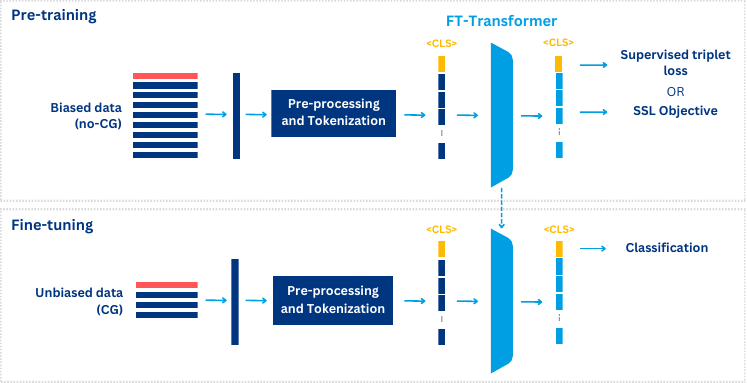}
    \centering
    \vspace{-5pt}
    \caption{Pre-training and fine-tuning framework for FT-Transformer utilizing unbiased and biased datasets.}
    \label{fig:framework}
\end{figure*}

\subsection{Deep Learning for Tabular Data}
Throughout the last 10-15 years, Deep Learning approaches explored in research studies and industrial applications have developed from shallow fully connected neural networks to deep models with sophisticated architectures containing billions of parameters. Nevertheless, the performance of neural networks on structured tabular data is still comparable to GBDTs \cite{gorishniy2021revisiting}. In the relevant literature, this is explained by the heterogeneity of tabular features and weaker correlations among them compared to co-occurring spatial, temporal, and semantic relationships in unstructured data sources (e.g., images or text) \cite{borisov2022deep}.

Recently, multiple architectures extending Transformer self-attention to tabular data have been proposed aiming to challenge and eventually overcome Gradient Boosting-based methods~\cite{YanCWC023, KossenBLGRG21, GorishniyRKSKB24, Hollmann0EH23,song2019autoint,huang2020tabtransformer,gorishniy2021revisiting}. In 2019, Song et al. \cite{song2019autoint} introduced the AutoInt architecture converting raw tabular features to embeddings that are passed through custom self-attention layers. TabTransformer \cite{huang2020tabtransformer}, introduced in 2020, exploits conventional multi-head self-attention layers \cite{vaswani2017attention} for categorical feature embeddings, whereas numerical features are processed via a separate MLP branch of the network. The FT-Transformer architecture \cite{gorishniy2021revisiting} tokenizes each numerical and categorical feature, appends a class token, and applies the classical Transformer self-attention layers to the tokens.
According to the original work, it outperforms the previously suggested Deep Learning models and even outperforms GBDTs on certain tasks. 

The emergence of tabular Transformers paves the way for adapting contemporary representation and transfer learning methods to tabular data. In particular, the SSL techniques can be used to pre-train the models on unlabeled datasets for further fine-tuning with available annotation. For example, the authors of TabTransformer \cite{huang2020tabtransformer} and FT-Transformer \cite{rubachev2022revisiting} adapted such well-known pre-training tasks as Masked Language Modelling (MLM) or Replaced Token Detection (RTD) to the tabular domain. Meanwhile, general frameworks, such as VIME \cite{zhang2020vime}, SCARF \cite{bahri2021scarf} and SAINT \cite{somepalli2021saint}, proposed pre-training for tabular DL models using reconstruction-based and contrastive learning methods. As a result, SSL methods applied to tabular data are especially advantageous compared to the GBDTs when addressing tasks where only a small proportion\footnote{Here, small should be read as "small w.r.t. all available data", which can still be relatively large in absolute numbers.} of data can be used for supervised training.
In addition to that, SSL enables general-purpose feature encoding which compresses information about structured tabular data.
Such representations of table rows can later be plugged into higher-level modules that model different sorts of relations between them, e.g. time-based sequences, but this area goes beyond our scope.

\vspace{-7pt}
\subsection{Fraud Detection}
\noindent\textbf{Fraud detection models.}
In the realm of e-commerce systems, the ability to effectively prevent fraudulent behaviours allows companies to create a safe and fair environment for their users. Thus, automatic fraud detection has gained a significant amount of attention in industrial applications and recent research literature. Whereas the simplest fraud detection algorithms can operate through a set of rule-based heuristics, more sophisticated systems additionally utilize machine learning to capture non-trivial fraudulent patterns in tabular data \cite{kou2004survey, ryman2018artificial}. Predominantly, fraud detection is addressed as a binary classification task by training classical Machine Learning, including Gradient Boosting, and Deep Learning methods in a supervised fashion with fully annotated datasets \cite{varmedja2019credit, taha2020intelligent, liu2022user, yu2022metarule}. Nevertheless, it is worth noting that some works formulate the problem as anomaly detection and utilize one-class and unsupervised methods for detecting fraudulent instances \cite{carcillo2021unsup, leevy2023investigating, xu2023deep}. 

We acknowledge that multiple studies in the literature \cite{cheng2020graph, lu2022graphebay, Ge2022compgraph, forough2021ensemble} address the problem of fraud detection using sequences of transactions and/or graph models. Nevertheless, in our case, we are focused on item-level fraud detection for two reasons. First, the existing data and model pipelines operate on instance-level data, processing datapoints in real time. Thus, for a fair comparison with GBDTs, we used the same structure of data in our experiments. Furthermore, more sophisticated Deep Learning-based approaches utilize raw tabular features describing user actions as items and nodes in sequences and graphs. Therefore, embeddings learnt for these individual rows can be used in future work for initializing states in sequential and graph-based models.

\noindent\textbf{Selection bias.}
The development of fraud detection systems presents various challenges that have been studied by researchers, such as highly imbalanced and shifting label distributions, annotation delays, and constantly evolving fraudulent patterns \cite{dal2017credit}. Nevertheless, yet another issue inevitably arising in real-world systems, namely selection bias, is often overlooked in research literature. In Machine Learning systems, selection bias can be defined as the discrepancy between training data and unseen data in production that occurs from a non-random selection of instances for model training \cite{Zadrozny2004}. Besides, as mentioned earlier in Section \ref{sec:intro}, certain domain-specific challenges, associated with getting data annotations, are another possible source of large discrepancies between training data and real transaction traffic in production. To address this issue, the Control Croup strategy (Figure \ref{fig:CG}), which holds out a small set of instances from the whole traffic for data labeling, can be utilized to get a representative distribution of data for model training with trustworthy annotations \cite{tax2021bookingfraud}.


Machine Learning algorithms trained under selection bias can benefit from domain adaptation techniques that aim to correct model training and predictions accounting for various shifts between unbiased and control datasets. Such techniques include but are not limited to sample-based techniques, feature-based methods, and approaches operating at inference time \cite{kouw2019review, farahani2021brief, boeken2023correcting}. In turn, this study addresses this problem in a slightly different manner. In particular, we explore various pre-training techniques applied to tabular Transformers to learn robust representations of row-level instances on all available data, biased and unbiased. Subsequently, we fine-tune the Transformers along with the discriminative classification models on Control Group data, which is not exposed to the bias introduced by the fraud detection system.

\vspace{-1pt}
\section{Methodology}
\subsection{Problem Definition}

Fraud Detection with tabular data is typically addressed as a binary classification problem. A data instance (e.g. a user action or a transaction) is represented by a feature vector $\boldsymbol{x}_i = (\boldsymbol{x}^{num}_i, \boldsymbol{x}^{cat}_i)$ that consists of $M$ numerical features $\boldsymbol{x}^{num}_i \in \mathbb{R}^M$ and $L$ categorical features $\boldsymbol{x}^{cat}_i =(x^{cat}_{i1}, \dots, x^{cat}_{iL})$, where $x^{cat}_{ij} \in \{1, 2, \dots\, C_j\}$ and $C_j$ is a cardinality of $j$-th categorical feature. Each annotated instance is associated with a binary label $y_i \in \{0, 1\}$. The datapoint is assigned with label $0$ if it is genuine and with label $1$ if it is fraudulent.

Let $\mathbb{V}=\{1, \dots\, C\}$, where $C=\sum_{j=1}^L C_j$, be a global vocabulary\footnote{
We use global vocabulary here and below in order to simplify notation. Categorical values can be easily mapped to the vocabulary entries given offsets $O_j=\sum_{k=1}^{j-1} C_k$ so that we have $v = x^{cat}_{\cdot j} + O_j$ for some $v\in\mathbb{V}$.
} for all categorical features.
Then, a fraud detection model is a function $h_{\theta}: \mathbb{R}^{M}\times\mathbb{V}^L \xrightarrow{} [0,1]$ which maps input features to the probability of the corresponding instance being fraudulent.
In deep learning, this function is typically composed of two distinct neural networks, a feature encoder and a classifier head. The former embeds potentially sparse\footnote{
Here, sparsity comes mainly from categorical features. Indeed, being represented with the one-hot encoding, they become sparse vectors in $\mathbb{R}^C$ which has very high dimension in case of high cardinalities.
} input features to some dense latent space $f_{\theta_e}: \mathbb{R}^{M}\times\mathbb{V}^L \xrightarrow{} \mathbb{R}^D$, whilst the latter actually performs binary classification in this space $g_{\theta_{clf}}: \mathbb{R}^D \xrightarrow{} [0, 1]$. The classifier is put on the top of the encoder making a well-known stacked architecture $h_{\theta}(\boldsymbol{x}_i) = g_{\theta_{clf}}(f_{\theta_e}(\boldsymbol{x}_i))$. In the supervised learning settings we fit this model to a labeled dataset $\mathcal{D} = \{(\boldsymbol{x}_i, y_i)\}_{i=1}^N$ using the maximum likelihood principle, which results in minimizing binary cross-entropy (BCE) objective between true labels and predicted probabilities
\begin{equation}
-\frac{1}{N}\sum_{i=1}^N y_i \log h_{\theta}(\boldsymbol{x}_i) + (1 - y_i) \log (1 - h_{\theta}(\boldsymbol{x}_i))\xrightarrow{}\min_{\theta}
\end{equation}
Thus, both model components are trained simultaneously in an end-to-end fashion.


Training fraud detection models on the non-randomly sampled set of datapoints available with annotations can lead to a sub-optimal performance in production, due to selection bias. Furthermore, making assumptions about the missing labels is risky because it could introduce additional bias. Thus, as discussed in Section \ref{sec:intro}, practitioners frequently use a so-called Control Group strategy, where unbiased dataset $\mathcal{D}_{cg} = \{(\boldsymbol{x}_i, y_i)\}_{i=1}^{N_{cg}}$ for supervised training is randomly sampled from the real traffic and annotated. The CG data contains only a small proportion of the whole set of instances, so the remaining data points are unused when training models in the end-to-end supervised fashion. In this work, we aim to leverage the representation learning ability of Deep Learning models in order to utilize a larger biased dataset $\mathcal{D}_{a} = \{(\boldsymbol{x}_i, y_i)\}_{i=1}^{N_a}$. Consisting of a significantly larger number of instances ($N_a >> N_{cg}$), this dataset is a non-random sample from the user traffic with a shifted distribution of features and labels. We propose to pre-train a Transformer-based feature encoder using two different approaches, Deep Metric Learning and Self-Supervised Learning, on the large dataset $\mathcal{D}_a$ for further fine-tuning on the unbiased Control Group $\mathcal{D}_{cg}$. The proposed framework is presented in Figure \ref{fig:framework}.


\subsection{Feature Encoder: FT-Transformer}
In the last years, the research on tabular representation learning has advanced significantly from using classical fully-connected neural networks trained from scratch to advanced architectures based on Transformer self-attention and pre-trained using SSL. Transformers, thanks to their self-attention mechanism, can model features in the context by learning high-order, non-linear feature interactions~\cite{song2019autoint}. In this study, we exploit a widely-used FT-Transformer \cite{gorishniy2021revisiting} to encode representations of incoming instances. This model has been selected because it allows the generation of compact representations for the whole row in a table (through the class token or global pooling) as well as for each input feature separately. Such architecture makes it possible to enable various pre-training techniques, including both the ones based on masked reconstruction and the ones based on metric or similarity learning, which we further describe in the subsequent sections.

\subsection{Supervised Deep Metric Learning}
\label{sec:meth_dml}

Deep Metric Learning (DML) has gained a significant amount of attention in the past years as a paradigm for learning latent representations. The main idea behind this paradigm is to learn representations in an embedding space reflecting the semantic similarity which, in the supervised setting, is defined using ground-truth labels \cite{roth2020revisiting}. That is typically achieved by specific loss functions that aim to align inputs from the same class in a latent space using distance- or similarity-based metrics. In practice, the alignment of embeddings is done for groups of instances, e.g. pairs, triplets \cite{schroff2015facenet}, or other sets \cite{chen2017beyond}, and specific loss functions are designed to minimize distances between the same label embeddings and, in turn, maximize distances between representations corresponding to different labels.

In the case of fraud detection, we define a DML model as a feature encoder $f_{\theta_e}: \mathbb{R}^{M}\times\mathbb{V}^L \xrightarrow{} \mathbb{R}^D$ mapping input instances to embeddings of size $D$. We propose utilizing a triplet loss function \cite{schroff2015facenet} to group fraudulent and genuine datapoints in a latent space. Specifically, a triplet of instances can be defined as $\{f_{\theta_e}(\boldsymbol{x}^a), f_{\theta_e}(\boldsymbol{x}^p), f_{\theta_e}(\boldsymbol{x}^n)\}$. In each triplet $f_{\theta_e}(\boldsymbol{x}^a)$ and $f_{\theta_e}(\boldsymbol{x}^p)$ are the anchor and positive embeddings corresponding to the same class label, also referred to as a positive pair, whereas $f_{\theta_e}(\boldsymbol{x}^n)$ is a negative instance belonging to another class. To build triplets, we use an online mining strategy that collects all positive pairs within a mini-batch and later combines them with all possible negatives. Moreover, motivated by findings in \cite{hermans2017defense}, we only consider semi-hard and hard triplets when computing the loss as follows:

\begin{equation}
\begin{aligned}
    L_{t} = \frac{1}{T - T_e}\sum_{i=1}^T max\Bigl\{0, ||f_{\theta_e}(\boldsymbol{x}^a_i) -  f_{\theta_e}(\boldsymbol{x}^p_i)||_2^2 \\  - ||f_{\theta_e}(\boldsymbol{x}^a_i) - f_{\theta_e}(\boldsymbol{x}^n_i)||_2^2 + \alpha \Bigr\},
\end{aligned}
\label{eq:triplet_loss_main}
\end{equation}
where $\alpha$ is a margin hyperparameter, $T$ is the number of all triplets in a mini-batch, and $T_e$ is the number of easy triplets. The latter is defined by the following inequality:

\begin{equation}
\begin{aligned}
    ||f_{\theta_e}(\boldsymbol{x}^a_i) -  f_{\theta_e}(\boldsymbol{x}^p_i)||_2^2 -||f_{\theta_e}(\boldsymbol{x}^a_i) - f_{\theta_e}(\boldsymbol{x}^n_i)||_2^2 + \alpha \le 0
\end{aligned}
\end{equation}

\subsection{Self-Supervised Learning}
\label{subsec:meth_ssl}
SSL has become very popular in the recent years thanks to the success of general-purpose pre-trained models for such domains as language and vision. These applications of SSL have popularised two major families of methods, reconstruction-based and metric-based~\cite{balestriero2023cookbook}. The former originates from Denoising AutoEncoders~\cite{vincent2008dae}, trying to reconstruct corrupted input features and optionally predict which ones have been corrupted. The latter comes from contrastive or metric learning objective and in the self-supervised setting aims to match original inputs with their corresponding corrupted versions in terms of some similarity or distance measure. This family is also applicable in the supervised setting, matching inputs that have the same label, which leads to the methods described previously in Section~\ref{sec:meth_dml}. Both families have been successfully applied to the tabular domain even without Transformer backbones~\cite{zhang2020vime, bahri2021scarf}. Recently, some researchers have also re-iterated on them in combination with tabular Transformer architectures~\cite{somepalli2021saint, rubachev2022revisiting}.

In this study, we have implemented and investigated only the first, reconstruction-based family\footnote{We decided not to study the SSL version of DML due to the nature of our target problem, detecting fraudulent items. Particularly, a large amount of labeled data enables the supervised version of this family. Selection bias makes training of the discriminative model more challenging, but does not affect supervised DML pre-training.} of SSL methods, closely following the work presented in~\cite{zhang2020vime}. More precisely, we employ a combined objective that consists of the weighted sum of reconstruction and mask predicting losses (Equations \ref{eq:loss_req} and \ref{eq:loss_mp}) given a set of original features $\boldsymbol{x}$ and their augmentations $\tilde{\boldsymbol{x}}$:
\begin{equation}
L=\gamma L_r + (1-\gamma) L_{mp},
\label{eq:loss_ssl_total}
\end{equation}
where $\gamma\in(0,1)$ is the weighting factor.
Augmentations randomly mix original and corrupted views, $\boldsymbol{x}$ and $\boldsymbol{x}_{corr}$, of the input feature vector using a binary mask $\boldsymbol{m}\in\{0,1\}^{M+L}$
\begin{equation}
\tilde{\boldsymbol{x}} = (1-\boldsymbol{m})\odot\boldsymbol{x}+\boldsymbol{m}\odot\boldsymbol{x}_{corr}.
\end{equation}

A corrupted view is created by shuffling data instances in the original table independently for each feature column, which is equivalent to sampling from their empirical marginal distributions.
Elements of mask vectors are drawn randomly from a Bernoulli distribution with a predefined probability called \textit{corruption rate}.

As we have two naturally different subsets of features, namely numerical and categorical, the reconstruction loss is the sum of the corresponding components:
\begin{equation}
L_r(\boldsymbol{x}, \hat{\boldsymbol{x}}) = L_{num}(\boldsymbol{x}^{num}, \hat{\boldsymbol{x}}^{num}) + L_{cat}(\boldsymbol{x}^{cat}, \hat{\boldsymbol{x}}^{cat})
\label{eq:loss_req}
\end{equation}

Minimizing the reconstruction loss is equivalent to maximizing log-likelihood, assuming certain underlying distributions for both types of features. In our case, these are a unit variance Gaussian and Multinomial respectively. So for numerical features, the reconstruction objective is effectively a mean squared error (MSE):
\begin{equation}
L_{num}(\boldsymbol{x}^{num}, \hat{\boldsymbol{x}}^{num}) = \frac{1}{2M} \sum_{i=1}^M \left(\hat{x}_{\cdot i}^{num} - x_{\cdot i}^{num}\right)^2,
\end{equation}
where $\hat{x}_{\cdot i}^{num}$ is a predicted value\footnote{
Here and below $(\cdot)$ in the subscript means an omitted instance-level index.
} of the $i$-th numerical feature,
whilst for categorical features it is a cross-entropy (CE):
\begin{equation}
L_{cat}(\boldsymbol{x}^{cat}, \hat{\boldsymbol{x}}^{cat}) = -\frac1L \sum_{i=1}^L \sum_{j=1}^{C_i} [x_{\cdot i}^{cat}=j] \log \frac{\exp{\hat{x}_{\cdot ij}^{cat}}}{\sum_{l=1}^{C_i}\exp{\hat{x}_{\cdot il}^{cat}}},
\end{equation}
where $\hat{x}_{\cdot ij}^{cat}$ are predicted logits of the $i$-th categorical feature and $[\cdot]$ is an indicator function.

An auxiliary objective $L_{mp}$ is aimed at predicting which features in the augmented input have been corrupted\footnote{In NLP it is also known as a replaced token detection (RTD).}.
So it is effectively a BCE with a mask vector as a target:
\begin{equation}
L_{mp}(\boldsymbol{m}, \hat{\boldsymbol{m}})=-\frac{1}{M+L}\sum_{i=1}^{M+L}m_{\cdot i}^{} \log \hat{m}_{\cdot i}^{} + (1 - m_{\cdot i}^{}) \log (1 - \hat{m}_{\cdot i}^{}),  
\label{eq:loss_mp}
\end{equation}
where $\hat{m}_{\cdot i}^{}$ is a predicted probability of the $i$-th feature being corrupted.

In order to predict values of numerical features, logits of categorical features and binary masks, we employ specifics of our Transformer backbone architecture.
More precisely, we benefit from contextualized modeling of the feature embeddings, so it is easy to perform feature-wise predictions using a set of shared SSL prediction heads.
Particularly, a Transformer encoder with no aggregation applied
$f_{\theta_e}:\mathbb{R}^M\times\mathbb{V}^L\xrightarrow{}\mathbb{R}^{D\times(M+L)}$ predicts one embedding vector for each input feature.
Then, we apply reconstruction and mask prediction heads to each such vector separately:
\begin{eqnarray}
    \hat{\boldsymbol{x}}^{num}&=&g_{\theta_n}(\boldsymbol{e}),
    \;g_{\theta_n}:\mathbb{R}^{D\times M}\xrightarrow{}\mathbb{R}^{M},\\
    \hat{\boldsymbol{x}}^{cat}&=&g_{\theta_c}(\boldsymbol{e}),
    \;g_{\theta_c}:\mathbb{R}^{D\times L}\xrightarrow{}\mathbb{R}^{L\times C},\\
    \hat{\boldsymbol{m}}&=&g_{\theta_m}(\boldsymbol{e}),
    \;g_{\theta_m}:\mathbb{R}^{D\times (M+L)}\xrightarrow{}[0,1]^{M+L},
\end{eqnarray}
where $\boldsymbol{e}=f_{\theta_e}(\tilde{\boldsymbol{x}})$ are embeddings of the augmented input.
SSL heads are shallow MLPs with a single output in the case of numerical features and masks\footnote{
A logistic sigmoid is applied to the output of the mask prediction head in order to fit it to the probability range.
} and with $C$ outputs in the case of categorical features.
\section{Offline Experimental Setup}



\subsection{Datasets}
\label{sec:datasets}
Experiments conducted in this work are based on internal confidential Booking.com datasets containing features generated from real user actions for certain fraud detection domain(s)\footref{booking_number_fn} from several months in 2022 and 2023. The train, validation, and test sets are obtained using time-based splitting, i.e. the validation and test splits are extracted from the months following the train set. Each instance in the dataset is defined by around one hundred numerical and a handful of binary hand-crafted features. For the former, we fill missing values with zeros whilst for the latter we treat them as a separate "unknown" category before building the vocabulary. As suggested in \cite{gorishniy2021revisiting}, quantile transformation is applied to the numerical features before feeding them to the FT-Transformer model. Given that fraud detection is a problem with severe data imbalance, we use the Average Precision (AP) to evaluate model performance. 

Below, we introduce the details of two datasets, namely CG and no-CG, used in this study:

\noindent The \textbf{biased (no-CG) dataset} contains all instances from the traffic for which labels are available, except the ones sampled for the Control Group. The bias in this dataset is caused by a specific selection procedure, which introduces a shift in both features and labels, meaning that this dataset does not represent production data. The total number of instances in this dataset has the order of tens of millions.

\noindent The \textbf{unbiased (CG) dataset}, obtained using the Control Group strategy, contains a smaller amount of instances randomly sampled from the user traffic with reconstructed labels. The number of instances in this dataset is between one and two orders of magnitude less than the no-CG dataset size.

While the distribution of labels in the CG and no-CG datasets varies, the proportion of fraudulent examples in both datasets is very low. This results in highly imbalanced datasets. It is worth noting that we do not apply any data balancing technique. However, to derive the decision threshold we use an optimization algorithm that balances the costs and benefits of positive and negative samples.\footnote{Because of confidentiality we cannot share more precise information.\label{booking_number_fn}}


\subsection{GBDT Baseline}
\label{sec:gbdts}

In this study, we use the LightGBM framework~\cite{ke2017lightgbm} as the GBDT baseline to compare against the proposed Deep Learning architectures. 
We analysed three strategies. For the first
we train only on the CG dataset. For the remaining two we
train on the merged CG and no-CG datasets, with and 
without \textit{selection ratio} (SR) weights, as suggested in~\cite{Zadrozny2004}. To be precise, we train
a classifier that predicts whether an instance of the
full population is annotated and weigh annotated training instances by the inverse of the (clipped) predicted probabilities.

For each strategy we optimized hyperparameters using 100 random search iterations with the uniform sampling strategy from the following search space:
\begin{itemize}
    \item learning rate $\in[0.02, 0.1]$,
    \item maximum depth $\in\{4,..,12\}$,
    \item sampling fraction of columns per tree $\in[0.1, 1.0]$,
    \item minimum number of instances per leaf node $2^l$, $l\in\{6,..,12\}$,
    \item bagging fraction $\in[0.1, 1.0]$,
    \item regularization parameters $\lambda,\alpha\in[0.0, 1.0]$.
\end{itemize}
All models were trained using up to 1000 trees in the ensemble applying early stopping with a patience of 25 rounds, based on the validation set performance, to prevent overfitting and ensure computational efficiency.

\subsection{Deep Learning Models}
\label{sec:dl}
\begin{table}[]
\caption{Base and large FT-Transformer hyperparameters and characteristics.}
\vspace{-16pt}
\scalebox{1.1}{
\begin{tabular}{ll|cc}
\multirow{4}{*}{Backbone}                            &                            & FT-base & FT-large \\ \hline
                            & token dimension       & 64      & 384      \\
                            & num. heads                 & 8       & 24       \\
                            & num. blocks                & 5       & 6        \\
                            & num. parameters (M) & 0.614   & 7.791    \\ \hline
\multirow{3}{*}{SSL}        & num. epochs           & 100     &    10      \\
                            & learning rate              &  0.001    &    0.001      \\
                            & eff. batch size              &  2048       &   4096      \\
                            \hline
\multirow{3}{*}{DML}        & num. epochs           & 100     &    10      \\
                            & learning rate              &  0.0001       &   0.0001      \\
                            & eff. batch size              &  4096       &   4096      \\
                            \hline
\multirow{2}{*}{Supervised} & num. epochs           & $\leq$100     &    $\leq$100      \\
                            & learning rate              &   0.0001      &   0.0001      \\
                            & eff. batch size              &  2048       &   2048      \\
\end{tabular}}
\vspace{-16pt}
\label{tab:transformer_hyperparams}
\end{table}

In the subsequent paragraphs, we describe the two backbones and employed training objectives. Table \ref{tab:transformer_hyperparams} summarizes the key hyperparameters of these models.

\vspace{2pt}
\noindent\textbf{Backbone architecture.}
In this paper, we used the official implementation\footnote{https://github.com/yandex-research/rtdl} of the FT-Transformer~\cite{gorishniy2021revisiting} as the backbone for all our experiments. In particular, two versions of this model, referred to as \textit{FT-base} and \textit{FT-large}, were implemented and evaluated. We trained all models for 100 epochs on the CG dataset and for 10 epochs when using the large no-CG dataset. We employed a standard AdamW optimizer with default parameters and the weight decay
of $10^{-5}$, unless specified otherwise.
We used constant learning rate without warm-up\footnote{
It is known~\cite{popel2018training_tf} that warm-up is important for Transformers to converge well. However, FT-Transformer employs layer pre-normalization instead of post-normalization, which can tolerate no warm-up~\cite{xiong2020prenorm}. In future experiments, we aim to explore the impact of different schedules on the convergence and performance of our models.
} to simplify our experiments. We saved model parameters based on the best value of validation loss.

\vspace{2pt}
\noindent\textbf{End-to-End Supervised Learning.}
The conventional supervised learning from scratch was performed by applying a linear layer with two outputs and softmax activations to the <CLS> token of the FT-Transformer model.
We used a learning rate of $0.0001$ and applied early stopping based on validation loss with a threshold of 0.0001 and patience of 20 epochs to prevent overfitting.


\vspace{2pt}
\noindent\textbf{Supervised Deep Metric Learning.}
The first family of the implemented pre-training methods, DML, is based on the triplet loss described in Section~\ref{sec:meth_dml}. We set margin $\alpha$ to 0.2 (Equation \ref{eq:triplet_loss_main}) and pre-trained both backbones without early stopping using the learning rate of 0.0001. 


\vspace{2pt}
\noindent\textbf{Self-Supervised Learning.}
The second family of the implemented pre-training methods relies on reconstruction-based SSL described in Section~\ref{subsec:meth_ssl}.
We assigned equal weights to both components of the pre-training objectives by setting $\gamma=0.5$ (Equation \ref{eq:loss_ssl_total}), as they were expected to have the same order of magnitude.
We employed a corruption rate of 0.4 to produce the augmented inputs.
Reconstruction and mask prediction heads are two-layer MLPs\footnote{
The two reconstruction heads share the first layer in order to reduce the number of parameters.
} with a hidden dimension that is twice as large as the token embedding dimension.
We set the weight decay to $2\cdot 10^{-5}$ and pre-trained both backbones without early stopping using the learning rate of 0.001.


\vspace{2pt}
\noindent\textbf{Supervised Fine-Tuning.}
The backbones pre-trained using DML and SSL were further fine-tuned for the fraud detection task in exactly the same fashion as the end-to-end supervised models.
The major difference was the initialization of the feature encoder weights from the pre-trained checkpoint instead of random.
The weights of the classification head were initialized randomly, as usual.

\section{Offline Evaluation}
Table~\ref{tab:baseline_comparison} summarizes the performance of the proposed models and baselines on the test set built upon the Control Group, which better represents a production-like distribution of data and labels. 

The first section of the table presents the test AP scores of three LightGBM baselines (Section \ref{sec:gbdts}). In particular, the first model was trained on CG data only whereas the second one utilized both CG and no-CG datasets. The third one additionally employed selection ratio weights, with the annotation classifier trained on the full dataset. The impact of selection bias is evident from the observed performance decline of over 2\% in the baselines trained on combined CG and no-CG data. Although this difference might not seem critical, it significantly affects business metrics and costs when scaled to the volume of user traffic.

\looseness-1
The second section demonstrates the results of various training settings for the base FT-Transformer. It includes supervised training from scratch on CG data and fine-tuning from feature encoders pre-trained using DML and SSL. Pre-training was performed on the CG dataset with extra fraudulent (positive) instances extracted from the no-CG dataset, which is referred to as \textit{CG + no-CG (+)} in the table. According to the table, FT-base, even without hyperparameter tuning, demonstrates performance comparable to thoroughly tuned LightGBM.
Remarkably, pre-training does not make a big difference in this setting.

The last section of the table shows the performance of the large FT-Transformer with and without the proposed pre-training strategies that use both CG and no-CG data. It is worth noticing that training FT-large from scratch on CG-only data fails to achieve baseline performance.
Presumably, there is not enough data to learn good representations.
However, pre-training it on a sufficiently large corpus (CG + no-CG) leads to superior performance compared to all other models.
SSL looks more promising than supervised DML in this context since it is less computationally expensive\footnote{Triplet loss in DML computes pairwise distances between all samples in a batch, which results in quadratic complexity. In turn, the SSL objective depends linearly on the batch size.} and does not require labels, allowing even larger-scale pre-training.

\begin{table}[!t]
\caption{Average Precision scores for GBDT baselines and FT-Transformer variants. The mean test AP scores and corresponding standard deviation values are reported for 10 runs with different random seeds.
}
\vspace{-6pt}
\scalebox{0.82}{
\begin{tabular}{llccccc}
\# &Approach                        & Backbone               & Pre-training & Training  & Test AP        \\ \hline
1&Sup.                     & LightGBM               &  -                                    &   CG        & 0.471 $\pm$ 0.003 \\
2&Sup.                     & LightGBM               &  -                               &   CG + no-CG        & 0.441 $\pm$ 0.013           \\
3&Sup.  + SR                   & LightGBM               & All data                                   &   CG + no-CG        & 0.442 $\pm$ 0.011          \\
\hline
4&Sup.             & FT-base  & -                                      &  CG        & 0.463 $\pm$ 0.013          \\ 
5&DML         & FT-base  & CG + no-CG (+)                   &  CG         & 0.469 $\pm$ 0.017 \\
6&SSL & FT-base  & CG + no-CG (+)                   & CG         &   0.469 $\pm$ 0.013         \\ \hline
7&Sup.              & FT-large & -                                     & CG          & 0.454 $\pm$ 0.009           \\
8&DML         & FT-large & CG + no-CG                                  & CG          & 0.488 $\pm$ 0.008           \\
9&SSL & FT-large & CG + no-CG                                 & CG & \textbf{0.491 $\pm$ 0.012}
\end{tabular}}

\label{tab:baseline_comparison}
\vspace{-6pt}
\end{table}

\section{Online Evaluation}
\noindent We conducted an online A/B experiment to further compare the performance of GBDTs with FT-Transformer for fraud detection at Booking.com. We followed the standard experiment protocol established within the scope of the production model. In the subsequent paragraphs, we describe the setup and the results of the experiment in as many details as possible, following internal company policies.

\vspace{2pt}
\looseness-1
\noindent\textbf{Evaluation metric.} To evaluate online performance we use an internal utility-based business metric that aggregates positive and negative effects of decisions made by the model on each data instance, rewarding correct ones and penalizing wrong ones. The metric has a high variance, which we overcome through Trigger Analysis~\cite{DengH15}, i.e. tracking only the ``triggered'' part of the traffic where the base and the variant models make different decisions. In our case, Trigger Analysis yields a notably higher statistical power.

\vspace{2pt}
\noindent\textbf{Base and variant models.} The base model is a LightGBM trained and hyper-parameter tuned as described in Section~\ref{sec:gbdts}, using SR weights (approach \#3 in Table~\ref{tab:baseline_comparison}). This approach has been shown to outperform approach \#1 in Table~\ref{tab:baseline_comparison} in previous A/B experiments and has already been used in production.
The variant model is the FT-Transformer trained with the Self-Supervised Learning strategy, as described in Section~\ref{sec:dl} (approach \#9 in Table~\ref{tab:baseline_comparison}). Both models were trained and validated using the same data splits. We used the most recent data available at start of the of online experiment for training both models, which is similar to the datasets used in the offline experiments in terms of size and label distribution. Selecting the decision threshold was also performed using the same algorithm for both models~(see Section~\ref{sec:datasets}).

\noindent\textbf{Training resources.} For training and hyperparameter tuning of the base model, we used a machine only with CPU resources. The variant model was trained on a machine with both GPU and CPU resources. For the same amount of data, and without performing any hyperparameter tuning for the variant, the training time is roughly six times higher for the variant.

\vspace{2pt}
\looseness-1
\noindent\textbf{A/B setup.} Both variant and base models were deployed in the same way to our internal machine learning serving platform. They worked in production simultaneously for a period of time long enough to result in  sufficient statistical power in the evaluation. To facilitate ``Trigger Analysis'', both models were evaluated in production for each fraud detection instance, though the model deciding on the action was selected at random. It is worth noting that even though the real-time latency for the variant is around 20 times higher than for the base, it still falls under the budget we have for responding to live requests.

\vspace{2pt}
\looseness-1
\noindent\textbf{Results.} The FT-Transformer variant outperforms the LightGBM baseline in terms of our business metric with a substantial margin. Specifically, we observe a 19.55\% uplift in the triggered part of the traffic. The improvement is statistically significant, using an unpaired t-test with the significance level $\alpha=0.01$. It is worth mentioning that the additional training costs are negligible in light of improvements in the business metric.\footref{booking_number_fn}
\vspace{8pt}
\section{Conclusions and Future Work}
In this paper, we explored the recent advancements in tabular representation learning for the challenging task of fraud detection. In particular, motivated by the problem of selection bias, we scaled multiple pre-training strategies for the Transformer-powered deep learning model to large amounts of unlabeled data. According to various offline evaluation scenarios, the SSL pre-training strategy confidently outperforms one of the industry-standard GBDTs as well as Transformers trained from scratch in a supervised manner. Moreover, we conducted online evaluation comparing FT-Transformer with LightGBM, our GBDT baseline, in an A/B experiment. We observed a statistically significant improvement over the baseline in terms of our internal utility-based business metric, which is inline with our offline findings.


Multiple research directions can be explored in the subsequent works on representation learning for the fraud detection domain. First, embeddings of raw features learnt with SSL can be plugged into more sophisticated architectures that model user actions as sequences and graphs or even trained end-to-end with them. Another promising idea is exploiting the latest cross-table learning frameworks to transfer knowledge between different tasks and domains within fraud detection systems and beyond.
\section*{Acknowledgement}
We would like to thank Maryam Moradbeigi for providing input and practical support to Bulat during his internship.
\newpage

\section*{GenAI Usage Disclosure}
Generative AI software tools were used to edit and improve the quality of the existing text in the same way we would use a typing assistant. They were also used in coding the same way we would use search engines and auto completion assistants. 
\bibliography{sample-base}
\end{document}